\documentclass[letterpaper, 10 pt, conference]{ieeeconf}  

\IEEEoverridecommandlockouts                              

\usepackage{epsfig} 

\usepackage{color}
\usepackage{authblk}
\usepackage{graphicx,hyperref}
\usepackage{amsmath}
\usepackage{amsfonts}
\usepackage{mathtools,amssymb,lipsum,caption,cuted}
\usepackage{subeqnarray}
\usepackage{multirow}
\usepackage{marvosym}
\usepackage{relsize}
\usepackage{siunitx}
\usepackage{lipsum}
\usepackage{nomencl}
\usepackage{caption}
\usepackage{soul}
\renewcommand\nomgroup[1]{%
  \item[\bfseries
  \ifstrequal{#1}{S}{List of Subscripts}{%
  \ifstrequal{#1}{P}{List of Symbols}{}%
  \ifstrequal{#1}{A}{List of Abbreviations}{}%
  }]%
}
\makenomenclature

\nomenclature[A]{\textbf{EMLA}}{Electromechanical linear actuator}
\nomenclature[A]{\textbf{EHLA}}{Electrohydraulic linear actuator}
\nomenclature[A]{\textbf{PMSM}}{Permanent magnet synchronous motor}
\nomenclature[A]{\textbf{OHM}}{Off-highway machine}
\nomenclature[A]{\textbf{BEV}}{Battery electric vehicle}
\nomenclature[A]{\textbf{ID}}{Inverse dynamics}
\nomenclature[A]{\textbf{TCP}}{Tool center point}
\nomenclature[A]{\textbf{ADC}}{Analog-to-digital converter}
\nomenclature[P]{$V_a$ $V_b$ $V_a$}{Three-phase electric motor voltage in (\SI{}{\volt})}
\nomenclature[P]{$V_d$ $V_q$}{Electric motor voltage in $dq$-axis (\SI{}{\volt})}
\nomenclature[P]{$i_d$}{Electric motor current in d-axis (\SI{}{\ampere})}
\nomenclature[P]{$i_q$}{Electric motor current in q-axis (\SI{}{\ampere})}
\nomenclature[P]{$R_s$}{Electric motor stator resistance (\SI{}{\ohm})}
\nomenclature[P]{$L_d$}{Electric motor inductance in d-axis (\SI{}{\henry})}
\nomenclature[P]{$L_q$}{Electric motor inductance in q-axis (\SI{}{\henry})}
\nomenclature[P]{$\boldsymbol{P}$}{Park transformation matrix}
\nomenclature[P]{$p$}{Number of pole pairs}
\nomenclature[P]{$\omega_m$}{PMSM angular velocity (\SI{}{\radian\per\sec})}
\nomenclature[P]{$\omega_e$}{Electric angular velocity (\SI{}{\radian\per\sec})}
\nomenclature[P]{$t_m$}{Final time (\SI{}{\sec})}
\nomenclature[P]{$t_0$}{Initial time (\SI{}{\sec})}
\nomenclature[P]{$\Phi_{PM}$}{Permanent magnet linkage flux in motor (\SI{}{\weber})}
\nomenclature[P]{$\rho$}{Screw mechanism lead (\SI{}{\meter})}
\nomenclature[P]{$G$}{Gear ratio ($G>1$)}
\nomenclature[P]{$\tau_m$}{Electric motor electromagnetic torque (\SI{}{\newton\cdot\meter})}
\nomenclature[P]{$\tau_m$}{Rotor friction (\SI{}{\newton\cdot\meter})}
\nomenclature[P]{$J_m$}{Rotor inertia (\si[inter-unit-product=\cdot]{\kilo\gram\meter\squared}
)}
\nomenclature[P]{$\ddot\theta_m$}{Electric motor angular acceleration (\si[per-mode=symbol]{\radian\per\second\squared})}
\nomenclature[P]{$f_{sm}$}{Force applied to the screw mechanism (\SI{}{\newton})}
\nomenclature[P]{$f_{x}$}{EMLA output force (\SI{}{\newton})}
\nomenclature[P]{$v_{x}$}{Linear velocity of the EMLA (\SI[per-mode=symbol]{}{\meter\per\second})}
\nomenclature[P]{$\dot{v}_{x}$}{Linear acceleration of the EMLA (\SI[per-mode=symbol]{}{\meter\per\second\squared})}
\nomenclature[P]{$M_{sm}$}{Screw mechanism mass (\SI{}{\kilo\gram})}
\nomenclature[P]{$b_{f}$}{Screw mechanism friction coefficient (\SI[per-mode=symbol]{}{\newton\cdot\second\per\meter})}
\nomenclature[P]{$cos(\phi)$}{Electric motor power factor}
\nomenclature[P]{$V_{LL}$}{Line-to-line voltage of 3-ph electric motor (\SI{}{\volt})}
\nomenclature[P]{$I_{LL}$}{Line-to-line current of 3-ph electric motor (\SI{}{\ampere})}
\nomenclature[P]{$\eta_{\text{EMLA}}$}{EMLA mechanism efficiency}
\nomenclature[P]{$L$}{First upper link length (\SI{}{\meter})}
\nomenclature[P]{$L_1$}{Second upper link length (\SI{}{\meter})}
\nomenclature[P]{$L_c$}{First lower link length (\SI{}{\meter})}
\nomenclature[P]{$L_{c0}$}{Second lower link length (\SI{}{\meter})}
\nomenclature[P]{$\theta \, \theta_1 \, \theta_2$}{Passive-joints angular positions (\SI{}{\radian})}
\nomenclature[P]{$q \, q_1 \, q_2$}{Internal trigonometric angles (\SI{}{\radian})}
\nomenclature[P]{$\psi \, \psi_1 \, \psi_2$}{Passive-joints angular offsets (\SI{}{\radian})}
\nomenclature[P]{$\Sigma_i$}{Spatial reference frame}
\nomenclature[P]{$\bld{G}$}{Homogeneous transformation matrix}
\nomenclature[P]{$\bld{\nu}$}{Twist vector}
\nomenclature[P]{$\dot{\bld{\nu}}$}{Spatial acceleration vector}
\nomenclature[P]{$\bld{B}$}{Matrix basis function}
\nomenclature[P]{$\bld{c}$}{Control points}
\nomenclature[P]{$\bld{\xi}$}{Particular decision variable}
\nomenclature[P]{$\bld{\Xi}$}{General decision variable}
\nomenclature[P]{$\bld{x}_r$}{Reference trajectory to track}
\nomenclature[P]{$\bld{J}$}{Manipulator Jacobian}
\nomenclature[P]{$\dot{\bld{J\,}}$}{Time derivative of the manipulator Jacobian}
\nomenclature[P]{$f$}{Cost function}

\graphicspath{ {Images/} } 

\newcommand{\real}{\mbox{\rm I$\!$R}}

\newcommand{\defeq}{\triangleq}
\newcommand{\bld}[1]{\mbox{\boldmath $#1$}} 

\usepackage[ruled,vlined,lined,linesnumbered,english]{algorithm2e}

\SetCommentSty{mycommfont}
\SetKwProg{Fn}{Function}{}{}

\begin{document}

\title{\LARGE \bf
	Energy-Cautious Designation of Kinematic Parameters for a Sustainable Parallel-Serial Heavy-Duty Manipulator Driven by Electromechanical Linear Actuator
}

\author{
    Alvaro Paz$^{*}$, Mohammad Bahari, and Jouni Mattila\\
    Faculty of Engineering and Natural Sciences, Tampere University, 33720, Finland\\
    *Email: alvaro.pazanaya@tuni.fi
    \thanks{This work was supported by the Business Finland partnership project “Future all-electric rough terrain autonomous mobile manipulators” (Grant \#2334/31/2022).}
}


\newpage
\thispagestyle{empty} 
    © 2024 IEEE. Personal use of this material is permitted. Permission from IEEE must be obtained for all other uses, including reprinting/republishing this material for advertising or promotional purposes, collecting new collected works for 
    resale or redistribution to servers or lists, or reuse of any copyrighted component of this work in other works. This 
    work has been submitted to the IEEE for possible publication. Copyright may be transferred without notice, after which this
    version may no longer be accessible.
\newpage 

\maketitle
\thispagestyle{empty}
\pagestyle{empty}

\begin{abstract}
Electrification, a key strategy in combating climate change, is transforming industries, and off-highway machines (OHM) will be next to transition from combustion engines and hydraulic actuation to sustainable fully electrified machines. Electromechanical linear actuators (EMLAs) offer superior efficiency, safety, and reduced maintenance, and they unlock vast potential for high-performance autonomous operations. However, a key challenge lies in optimizing the kinematic parameters of OHMs' on-board manipulators for EMLA integration to exploit the full capabilities of actuation systems and maximize their performance.
This work addresses this challenge by delving into the structural optimization of a prevalent closed kinematic chain configuration commonly employed in OHM manipulators. Our approach aims to retain the manipulator's existing capabilities while reducing its energy expenditure, paving the way for a greener future in industrial automation, one in which sustainable and high-performing robotized OHMs can evolve. The feasibility of our methodology is validated through simulation results obtained on a commercially available parallel-serial heavy-duty manipulator mounted on a battery electric vehicle. The results demonstrate the efficacy of our approach in modifying kinematic parameters to facilitate the replacement of conventional hydraulic actuators with EMLAs, all while minimizing the overall energy consumption of the system.
\end{abstract}

\printnomenclature

\section{Introduction}
\subsection{Background and Context}
\begin{figure}[h!] 
	\centering
	\includegraphics[trim={0.0cm 0.0cm 0.0cm 0.0cm},clip,width=8.7cm]{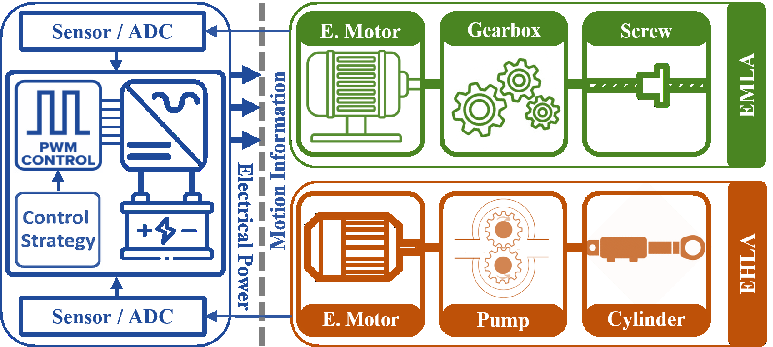}
	\vspace*{0cm}
	\caption{{The composition of an EMLA and an HLA}. }
	\label{fig:EMLA&EHLA}
\end{figure}

The transition to zero-emission solutions is reshaping the logistics and transportation industries, particularly in urban environments, where stringent regulations are driving the move away from internal combustion engines (ICEs) \cite{yang2022sustainable}. A prime example of this trend is Stockholm, Sweden, which is set to implement a stringent ban on diesel engines, including trucks with cranes, by the end of 2024. This initiative is part of a broader effort to establish low-emission zones across Europe, driven by urban vehicle access regulations (UVARs) intended to improve air quality and reduce congestion \cite{lopez2018urban}. Mirroring this shift towards zero-emission solutions in urban settings, the field of off-highway machines (OHMs) is also undergoing significant changes. Driven by stringent environmental regulations, such as the Paris Agreement of 2015 \cite{agreement2015paris}, and the push for reduced carbon footprints, the electrification of OHMs is transforming heavy-duty machinery, promising enhanced automation and sustainability, as well as greener future in this industry \cite{beltrami2021electrification}. To address the challenges and opportunities presented by the trend of electrification, technological advancements in actuator systems are essential \cite{9790066}. Currently, hydraulic linear actuators and pumps are employed in truck-mounted material handling cranes, hook and tail lifts. Hiab has already introduced an innovative solution in which the hydraulic pump is controlled by a permanent magnet synchronous motor (PMSM), marking a significant step toward electrification and sustainability in OHMs. While this represents progress, hydraulic actuators still present limitations. Although hydraulic actuators provide superior power density, they are constrained by issues such as energy inefficiency and susceptibility to leakages. Moreover, they require an additional energy conversion from electric power to hydraulic flow, which is then converted to mechanical motion. As a result, this has led to further innovation in the form of electromechanical linear actuators (EMLAs). EMLAs are an emerging technology also for heavy-duty linear actuation mechanisms, with required output forces ranging from 100 kN to 500 kN; compared to hydraulics, they offer superior efficiency and have lower maintenance requirements \cite{4211345}. EMLAs could enable more effective use of the limited stored energy in battery electric vehicles (BEV), offering a direct drive system that avoids the energy conversion losses inherent in hydraulic systems and a potentially extended range despite their tasks involving heavy-weight cargo lifting and handling \cite{10199841}. The compositions of both EMLAs and the electro-hydraulic linear actuator (EHLA) introduced by Hiab are illustrated in Fig. \ref{fig:EMLA&EHLA}. As urban areas implement ambitious emission reduction plans, integrating EMLAs into machinery aligns more closely with their environmental goals, offering a more sustainable and efficient solution for high-performance and autonomous operations.
However, achieving the benefits of EMLA hinges on successfully integrating EMLAs into existing OHMs' manipulator structures, ensuring not only that the new actuators seamlessly fulfill the task requirements previously handled by the conventional actuation mechanism of the manipulator but also that EMLAs' efficiencies are at the peak and actuation mechanism energy consumption is at lowest to conduct the tasks. Accordingly, optimizing the kinematic parameters of manipulators to find the optimal structure is crucial to implement these electric actuators.
Structural optimization has emerged as a powerful tool in manipulator design \cite{9043956}, offering a systematic approach to identifying the most efficient configuration within a vast design space. This field has garnered significant research interest in recent years, leading to the development of robust computer-aided optimization algorithms. Significant results have been reported in the structural design of heavy-duty robots, where high-performance numerical toolboxes (e.g., ANSYS Workbench and ADAMS software) have been employed to provide a representation of rigid multibody dynamics \cite{9043956,9620175} while the objective functions are targeted mainly to optimize structural deformation, harmonic response or mechanical stress. Such characteristics are crucial to study when considering heavy-duty machinery. In addition, parameter optimization design has been implemented to find spring values in a loaded balance system to minimize the consumption of joint drive \cite{liu2014} in a 5-DoF heavy-duty manipulator.

In the context of manipulator design, optimization focuses on defining the optimal geometric parameters to ensure effective performance across the desired range of motion. Because actuators play a critical role in the manipulator's lifting capacity, structural optimization aims to minimize peak actuator force within the operational range. This translates to a reduction in the required actuator capacity, leading to a more efficient and potentially lighter manipulator design. Given the vast design space explored through structural optimization, achieving mathematically optimal solutions can be computationally expensive. While achieving mathematically optimal solutions for manipulator design is theoretically possible through optimization algorithms or analytical methods, practical considerations often necessitate alternative approaches. Shoup \cite{shoup1980practical} utilized the golden section search for the min-max force problem. However, for real-world manipulator models, computationally efficient methods, such as the sequential quadratic programming (SQP) method implemented in MATLAB's $f_{minmax}$ function, are typically employed \cite{brayton1979new}. These methods may require modifications, such as incorporating exact merit functions alongside the Brayton \cite{brayton1979new} and Beiner \cite{beiner1995min} merit function.
%
%
\subsection{Paper Contributions}
\begin{figure}[h!] 
	\centering
	\includegraphics[trim={0cm 0cm 0cm 0.0cm},clip,width=9cm]{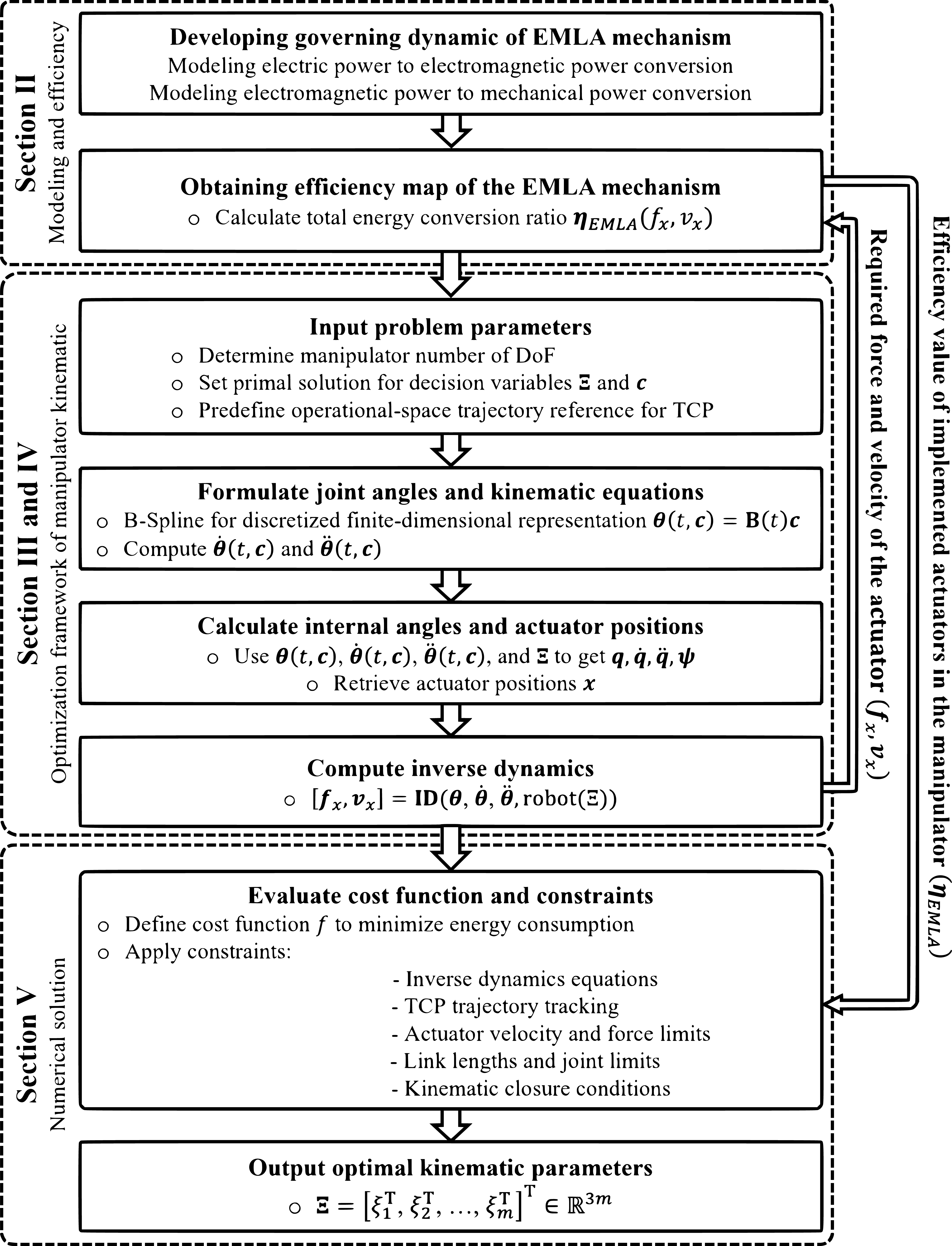}
	\vspace*{0cm}
	\caption{Procedural flowchart illustrating the methodology for energy-cautious manipulator kinematic parameter optimization.}
	\label{fig:Flowchart}
\end{figure}

This work addresses the critical challenge of integrating EMLAs into existing heavy-duty manipulator structures for sustainable OHM design. The key contributions of this paper are itemized as follows:
\begin{itemize}[leftmargin=0.35cm]
    \item Developed a mathematical model of EMLA mechanism and subsequently computed efficiency maps as a function of force and linear velocity at the load side.
    \item Introduced a high-fidelity analytical model of full parallel-serial manipulator dynamics, surpassing limitations of previous studies (e.g., \cite{9043956}, \cite{9620175}, and \cite{liu2014}) 
    \item Rather than analyze the structural stress or harmonic response, we developed an energy-centric optimization framework prioritizing energy consumption minimization through EMLA efficiency maximization to enhance the performance and sustainability of electrified OHM.
    \item Developed elegant analytical solutions for solving optimal problems when considering particular examples of robots and constraints \cite{beiner1995min}. However, our solution includes the full nonlinear dynamics model of the manipulator and its nonlinear constraints; thus, by transcribing the optimal problem, a discretized version is generated than can be solved with nonlinear numerical solvers.
    \item Proposed an optimization method to determine optimal kinematic parameters for a prevalent closed-kinematic chain manipulator configuration, enabling efficient EMLA integration and unlocking the potential of electrification while maintaining economic viability due to the reduced energy consumption of the actuation mechanism.
    \item Demonstrated the potential of this research to significantly reduce the overall environmental impact of transportation and logistics by facilitating the transition from ICEs to clean electric EMLAs in heavy-duty BEVs.
\end{itemize}
Fig. \ref{fig:Flowchart} visualizes the organization and methodology of the paper for energy-cautious manipulator kinematic parameter optimization.
\section{EMLA Mathematical Model}
\begin{figure}[bp] 
	\centering
	\includegraphics[trim={0.0cm 0.0cm 0.0cm 0.0cm},clip,width=8.7cm]{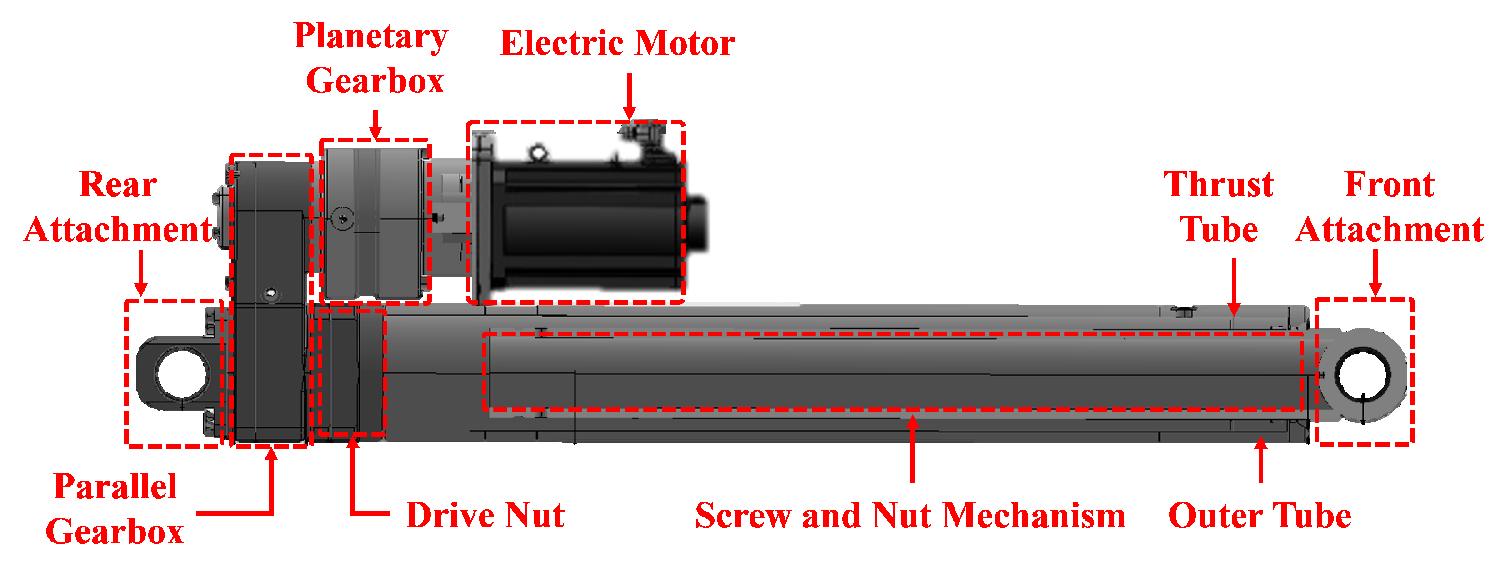}
	\vspace*{0cm}
	\caption{{The structure of an EMLA}. }
	\label{fig:EMLA_Structure}
\end{figure}

\begin{figure}[tbp] 
	\centering
	\includegraphics[trim={0cm 0cm 0cm 0.0cm},clip,width=8.5cm]{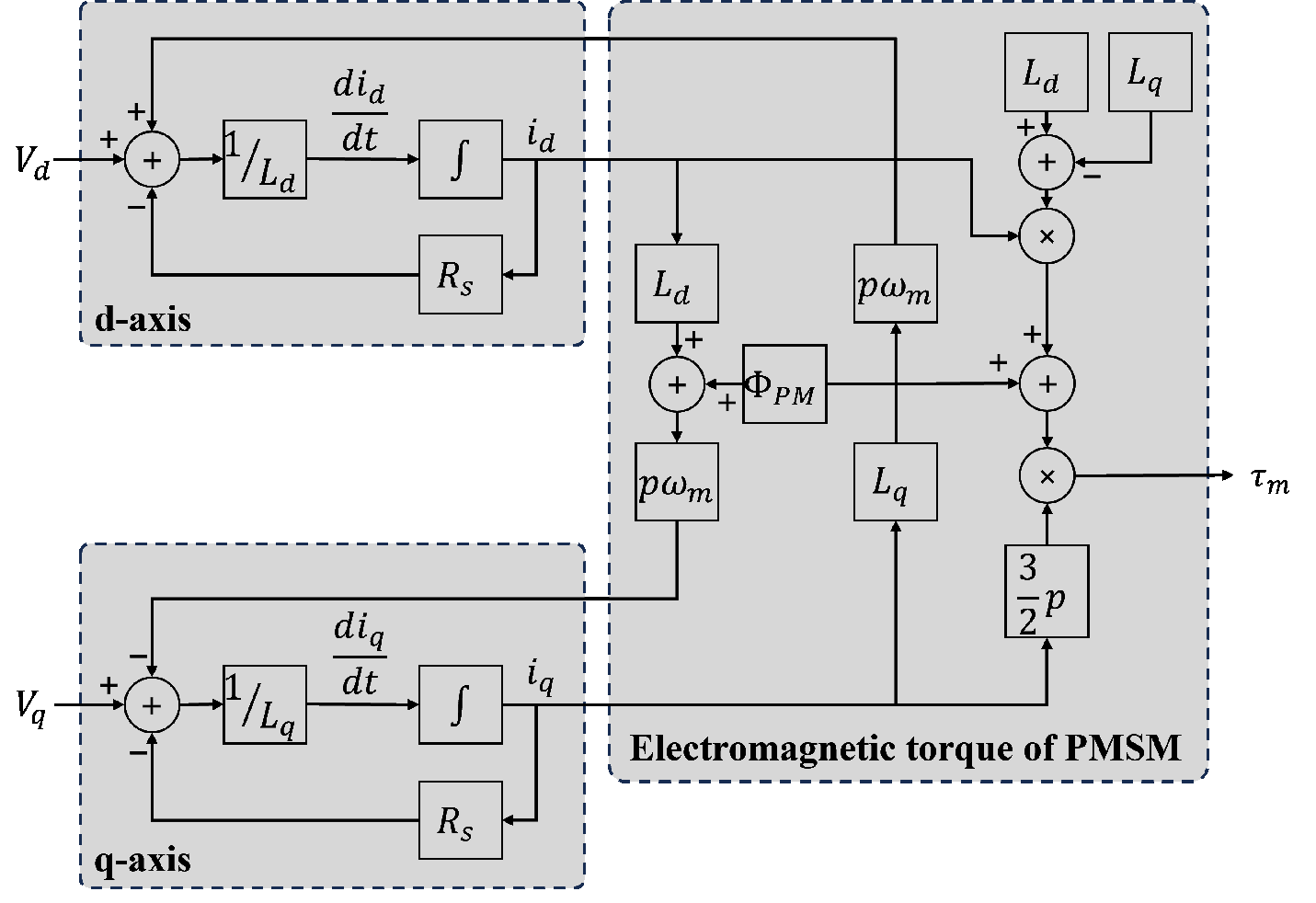}
	\vspace*{0cm}
	\caption{Equivalent model of a PMSM in the d-q reference frame}
	\label{fig:Equivalent}
\end{figure}

\begin{figure}[tbp] 
	\centering
	\includegraphics[trim={0cm 0cm 0cm 0.0cm},clip,width=9cm]{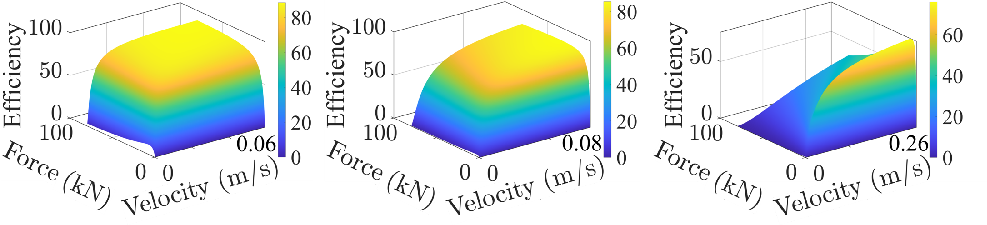}
	\vspace*{0cm}
	\caption{Efficiency of EMLAs as the function of force and linear velocity at load side}
	\label{fig:Eff}
\end{figure}

In this section, we investigate the mathematical model of an EMLA equipped with a PMSM, aiming to generate force ($f_x$) and linear velocity ($v_x$) at the load side of the actuator. The fundamental elements of an EMLA include an electric motor, gearbox, and screw mechanism, where the former provides torque and angular speed, the gearbox trades speed for increased torque, and the screw mechanism converts rotational motion into linear movement. This movement's force and velocity are functions of the gear ratio and screw's lead \cite{cao2011overview}. The main structural components of an EMLA is illustrated in Fig. \ref{fig:EMLA_Structure}. The power transmission sequence of the EMLA commences with (a) electric power into electromagnetic power, as described in \eqref{equation:Park_Transformation}-\eqref{eq:dq0voltage}, where subscripts $d$ and $q$ indicate the parameters in the $dq$ frame; (b) electromagnetic power into mechanical rotation, as explained in \eqref{eq:torque_torque}, where subscripts $m$, $sm$, and $f$ denote the parameters related to the electric motor, screw mechanism, and friction, respectively; and (c) rotational movement into linear movement, as expressed in \eqref{eq:torque_force}, where subscript $x$ shows the parameters at the load side. Power conversion occurs within the rotating $dq$ reference frame, employing Park transformations ($\boldsymbol{P}$) to decouple the three-phase system into orthogonal components, as detailed in \eqref{equation:Park_Transformation} and \eqref{equation:abc_to_dq}. In addition, the voltages and electromagnetic torque generated by the PMSM as a function of stator currents and permanent magnet flux are obtained as \eqref{eq:dq0voltage}. In the $dq$-axis, an equivalent model of the PMSM is illustrated in Fig. \ref{fig:Equivalent} \cite{6924771}.
\begin{equation}
\small
\hspace{-0.1cm} \boldsymbol{P} = \left[\begin{array}{ccc}
cos(\omega_e t)  & cos(\omega_e t - \frac{2 \pi}{3})  & cos(\omega_e t + \frac{2 \pi}{3}) \\
-sin(\omega_e t) & -sin(\omega_e t - \frac{2 \pi}{3}) & -sin(\omega_e t + \frac{2 \pi}{3}) \\
\frac{1}{2}    & \frac{1}{2}                      & \frac{1}{2}
\end{array}\right]
\label{equation:Park_Transformation}
\end{equation}
\begin{equation}
\left[\begin{array}{l}
V_d \\
V_q \\
V_0
\end{array}\right]
=\frac{2 \boldsymbol{P}}{3} 
\left[\begin{array}{l}
V_{\mathrm{a}} \\
V_{\mathrm{b}} \\
V_{\mathrm{c}}
\end{array}\right]
\label{equation:abc_to_dq}
\end{equation}
\begin{equation}
\left\{
\begin{aligned}
V_d &= R_s i_d + L_d \frac{d i_d}{dt} - p \omega_m L_q i_q \\
V_q &= R_s i_q + L_q \frac{d i_q}{dt} + p \omega_m (L_d i_d+ \Phi_{P\!M})\\
\tau_{m}&=\frac{3}{2} p i_q\left[\Phi_{P\!M}+\left(L_d-L_q\right) i_d\right]
\end{aligned}
\right.
\label{eq:dq0voltage}
\end{equation}
\begin{equation}
\tau_m - \tau_{f} = J_m \ddot\theta_m  +  \tau_{sm}
\label{eq:torque_torque}
\end{equation}
\begin{equation}
\left\{
\begin{aligned}
\tau_{sm} &= \frac{\rho}{2 \pi G} f_{sm}\\
f_{x} &= f_{sm} - M_{sm} \dot{v}_x - b_{f} v_x 
\end{aligned}
\right.
\label{eq:torque_force}
\end{equation}
The overall efficiency of the EMLA system is determined by the ratio of mechanical power output to electrical power input. As expressed in \eqref{system_efficiency}, the system efficiency, $\eta_{\text{EMLA}}$, is a function of time, load force $f_{x}$, and load velocity $v_{x}$. Fig. \ref{fig:Eff} illustrates the efficiencies of the EMLAs, which will be integrated as joint drives for the heavy-duty manipulator.
\begin{equation}
\eta_{\text{EMLA}}(f_{x}, v_{x}, t) = \frac{f_{x} (t) \cdot v_{x}(t)}{\sqrt{3} V_{LL}(t) I_{LL}(t) \cos {\phi}}
\label{system_efficiency} 
\end{equation}
\section{Structure of Parallel Mechanism}
\begin{figure}[h!] 
	\centering
	\includegraphics[trim={0cm 0cm 0cm 0.0cm},clip,width=3.75cm]{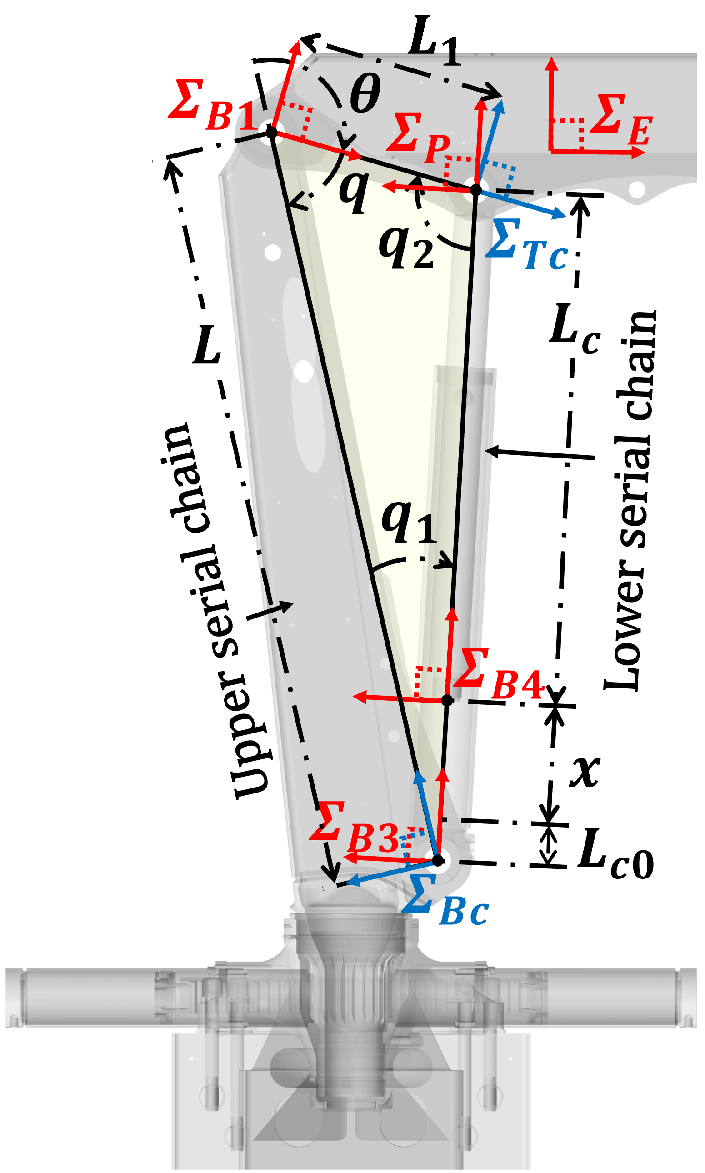}
	\vspace*{0cm}
	\caption{The parallel mechanism composed of three passive rotational joints and one linearly actuated joint. It can be kinematically analyzed as two separated serial chains, upper and lower, which meet in the reference frames $\Sigma_{Bc}$ and $\Sigma_{Tc}$. Each of the rotational joints has an associated internal angle, an offset angle, and a joint configuration angle. Such angles and their time derivatives are essential to solving the mechanism trigonometry, kinematics, and dynamics.}
	\label{fig:parallel}
\end{figure}

Heavy-duty parallel-serial manipulators are composed of prismatic, rotational and parallel mechanisms. Geometrically, this type of parallel mechanism under study is a one-degree of freedom (DoF) closed kinematic chain of four links with lengths $L, \, L_1, \, L_c, \, \text{and} \, L_{c0} > 0$ as shown in Fig. \ref{fig:parallel}. It is articulated by three passive rotational joints with angles $\theta$, $\theta_1$, and $\theta_2$ and one actuated prismatic joint with linear position $x$.
Such passive joints are placed at the origin of frames $\Sigma_{B1}$, $\Sigma_{B3}$, and $\Sigma_{T2}$ and the actuated one at $\Sigma_{B4}$.
%
%
Kinematically, this parallel mechanism can be studied as two serial kinematic chains restricted by holonomic constraints. Let us now define the following reference-frame sequences for each serial kinematic chain, which we call upper and lower. For the upper serial kinematic chain, the reference frames are $\Sigma_{B0}$, $\Sigma_{B1}$, $\Sigma_{T1}$, and $\Sigma_{E}$, and for the lower one, $\Sigma_{B2}$, $\Sigma_{B3}$, $\Sigma_{B4}$, $\Sigma_{P}$, and $\Sigma_{T2}$. At the beginning and end of the closed kinematic chain, general reference frames can be placed, such as
\begin{equation}
	\Sigma_{Bc}=\Sigma_{B0}=\Sigma_{B2}
		\qquad \mbox{and} \qquad \Sigma_{Tc}=\Sigma_{T1}=\Sigma_{T2}
		\label{eq:frames}
\end{equation}

As shown in Fig. \ref{fig:parallel}, the origins of the reference frames $\Sigma_{B1}$, $\Sigma_{Bc}$, and $\Sigma_{Tc}$ are the vertices of a triangle parameterized by link lengths and joint configurations. The inner angles of this triangle are denoted by $q$, $q_1$, and $q_2$, which are angles around the screw axis of the motion of the reference frames $\Sigma_{B1}$, $\Sigma_{Bc}$, and $\Sigma_{Tc}$, respectively. Such inner angles hold the triangle property
\begin{equation}
	q + q_1 + q_2 \ = \ -\pi
		\label{eq:triangle}
\end{equation}
and they can be computed once the link lengths and piston position $x\geq0$ are given using the expressions \cite{zhu2010virtual}
\begin{subequations}
	\begin{eqnarray}
		q & = & -\!\arccos\!\left(\! \tfrac{(x+x_{0})^{2}-L^{2}-L_{1}^{2}}{-2 L L_{1} } \right)  \label{eq:qus_a} \\
		q_{1} & = & -\!\arccos\!\left(\! \tfrac{L_{1}^{2}-(x+x_{0})^{2}-L^{2}}{-2(x+x_{0})L} \right) \\
		q_{2} & = & -\!\arccos\!\left(\! \tfrac{L^{2}-(x+x_{0})^{2} - L_{1}^{2}}{-2(x+x_{0})L_{1}} \right)
	\end{eqnarray}
 \label{eq:qus}
\end{subequations}
where $x_{0}\!=\!L_{c}\!+\!L_{c0}$.

Therefore, the passive joint angles can be calculated by
\begin{equation}
	\theta \ = \ q + \psi, \quad \theta_1 \ = \ q_1 + \psi_1, \quad \theta_2 \ = \ q_2 + \psi_2
 \label{eq:thetas}
\end{equation}
where $\psi$, $\psi_1$, and $\psi_2$ are offset angles that are constant scalar functions with respect to $x$ but not with respect to link length.

The geometric closure of the parallel mechanism is forced through holonomic constraints by the following expressions for internal angular velocities $\dot{q}$, $\dot{q}_1$, and $\dot{q}_2$ as linear functions of the linear velocity of the actuator $\dot{x}$.
\begin{equation}
	\dot{q} \ = \ k_1 \dot{x} \qquad\quad \dot{q}_{1} \ = \ k_2 \dot{x} \qquad\quad \dot{q}_{2} \ = \ k_3 \dot{x}
	\label{eq:dq__}
\end{equation}
where
\begin{subequations}
\begin{eqnarray}
	\hspace*{-1cm}k_1 & = & -\left( x+x_{0} \right) / \left(\!LL_{1}\!\sin q\right) \\
	\hspace*{-1cm}k_2 & = & -\!\left(x\!+\!x_{0} \!-\! L\!\cos q_{1}\right)/(x\!+\!x_{0})L\sin q_{1} \\
	\hspace*{-1cm}k_3 & = & -\!\left(x\!+\!x_{0} \!-\! L_{1}\!\cos q_{2}\right)/(x\!+\!x_{0})L_{1}\sin q_{2}
\end{eqnarray}
\label{eq:k1_3}
\end{subequations}
with $q$, $q_{1}$, and $q_{2} \neq 0$. As well, internal angular accelerations $\ddot{q}$, $\ddot{q}_1$, and $\ddot{q}_2$ are found by deriving (\ref{eq:dq__}) with respect to time as
\begin{equation}
	\ddot{q} = \dot{k}_1 \dot{x} + k_1 \ddot{x} \qquad \ddot{q}_{1} = \dot{k}_2 \dot{x} + k_2 \ddot{x} \qquad \ddot{q}_{2} = \dot{k}_3 \dot{x} + k_3 \ddot{x}
	\label{eq:04}
\end{equation}
where $\dot{k}_1$, $\dot{k}_2$, and $\dot{k}_3$ are the time derivatives of (\ref{eq:k1_3}) and $\ddot{x}$ is the actuator's linear acceleration.

In addition, the angular velocities of passive joints $\dot{\theta}$, $\dot{\theta}_1$, and $\dot{\theta}_2$ are obtained by deriving (\ref{eq:thetas}) with respect to time as
\begin{equation}
	\dot{\theta} \ = \ \dot{q}, \qquad \dot{\theta}_1 \ = \ \dot{q}_1, \qquad \dot{\theta}_2 \ = \ \dot{q}_2
 \label{eq:Dthetas}
\end{equation}
and their angular accelerations as
\begin{equation}
	\ddot{\theta} \ = \ \ddot{q}, \qquad \ddot{\theta}_1 \ = \ \ddot{q}_1, \qquad \ddot{\theta}_2 \ = \ \ddot{q}_2
 \label{eq:DDthetas}
\end{equation}

By adopting a screw theory notation \cite{lynch2017modern}, from definition (\ref{eq:frames}), it is assumed that reference frames $\Sigma_{B0}$ and $\Sigma_{B2}$ have the same transformation matrices $\bld{G}_{Bc}=\bld{G}_{B0}=\bld{G}_{B2} \in SE(3)$, twist vectors $\bld{\nu}_{\!_{\!Bc}}=\bld{\nu}_{\!_{\!B0}}=\bld{\nu}_{\!_{\!B2}} \in se(3)$, and spatial accelerations $\dot{\bld{\nu}}_{\!_{\!Bc}}=\dot{\bld{\nu}}_{\!_{\!B0}}=\dot{\bld{\nu}}_{\!_{\!B2}} \in se(3)$. Likewise, reference frames $\Sigma_{T1}$ and $\Sigma_{T2}$ have the same kinematic features as reference frame $\Sigma_{Tc}$. For more kinematic details using Lie groups and their algebras, see \cite{paz2024analytical}.

\section{Kinematic Parameters Optimization Problem}
The optimization problem can be formulated as follows. Given an 
$n$-DoF heavy-duty parallel-serial manipulator and a predefined operational-space trajectory reference for its TCP, find the optimal values for the link lengths $L$, $L_{c0}$, and $L_c$, for each of the parallel mechanisms, that minimizes the EMLA's energy expenditure while respecting robot constraints such as joint limits, force-and-velocity boundaries in the linear actuators, TCP reference trajectory tracking, boundaries for link lengths, and trigonometric closure conditions for closed chains consistency. In other words, the robot must be forced to perform the same TCP trajectory reference while the numerical optimizer finds the best link lengths that minimizes the energy consumption of the EMLAs while respecting structural restrictions.

Let us now define the decision variable vector for a single parallel mechanism (see Fig. \ref{fig:parallel}) as a stack of lengths
\begin{equation}
	\bld{\xi} \ = \ \begin{bmatrix}
		L & L_{c0} & L_c
	\end{bmatrix}^{\top} \ \in \ \real^{3}
\end{equation}
Thus, according to (\ref{eq:qus}) and (\ref{eq:thetas}), the internal angles $q_{1}(\bld{\xi})$ and $q_{2}(\bld{\xi})$, the offset angles $\psi_{1}(\bld{\xi})$ and $\psi_{2}(\bld{\xi})$, and the passive joint angles $\theta_{1}(\bld{\xi})$ and $\theta_{2}(\bld{\xi})$ become functions of $\bld{\xi}$.

Let us now adopt a BSpline parameterization of the configurational space trajectory \cite{Bib:Lee} to obtain a discretized finite-dimensional representation of our problem. For this a finite set of collocation points $\mathcal{T}$ transcribes the time as
\begin{equation}
\mathcal{T} \ \defeq \ \{t_0 \ \cdots \ t_k \ \cdots \ t_M\}
\end{equation}
where, $k$ is the iterator, and $M$ is the number of partitions.

The  vector $\bld{\theta}(t,\bld{c})\in\real^{n}$ that contains only the $\theta$ values of each parallel mechanism (i.e., first element of (\ref{eq:thetas})) and the joint positions of single-joint mechanisms can be defined by the matrix BSpline form
\begin{eqnarray}
\bld{\theta}(t,\bld{c}) & = & \bld{B}(t)\bld{c}
\label{eq:matrix_basis}
\end{eqnarray}
where $\bld{B}(t)\in\real^{n\times nN}$ is the matrix form of basis functions and $N$ is the number of control points stacked in $\bld{c}\in\real^{nN}$.

By time deriving the last equation twice, the following expressions for velocity and acceleration are obtained \cite{paz2019practical}
\begin{eqnarray}
\dot{\bld{\theta}}(t,\bld{c}) & = & \dot{\bld{B}}(t)\bld{c} \label{eq:matrix_basis_d} \\
\ddot{\bld{\theta}}(t,\bld{c}) & = & \ddot{\bld{B}}(t)\bld{c}
\label{eq:matrix_basis_dd}
\end{eqnarray}

Using $\theta$ from (\ref{eq:matrix_basis}) into (\ref{eq:thetas}) and (\ref{eq:qus_a}), the actuator position $x$ is retrieved, and then $x$ and $\bld{\xi}$ are used to compute all remaining internal angles and passive joint angles. Additionally, all angles and $x$ in the parallel mechanism become functions of the control points $\bld{c}$ and link lenghts $\bld{\xi}$.

The inverse dynamics of parallel-serial manipulators can be expressed by the function \cite{petrovic2022mathematical}
\begin{equation}
	\left[ \bld{f}_{\!x}, \ \bld{v}_{\!x} \right] \ = \  \bld{ID}\left( \bld{\theta}, \dot{\bld{\theta}}, \ddot{\bld{\theta}}, robot(\bld{\Xi}) \right)
 \label{eq:dynamics}
\end{equation}
where $\bld{f}_{\!x}$ and $\bld{v}_{\!x}$ are the actuator forces and velocities; $\bld{\theta}$, $\dot{\bld{\theta}}$, and $\ddot{\bld{\theta}}$ are functions of $\bld{c}$; $robot(\cdot)$ is the structure that contains all topological, kinematic, and inertial information of the robot and is a function of
\begin{equation}
\bld{\Xi} \ \defeq \ [ \bld{\xi}_{1}^{\top} \ \bld{\xi}_{2}^{\top} \ \cdots \ \bld{\xi}_{m}^{\top} ]^{\top} \in\real^{3m}
\end{equation}
which contains link lengths of all parallel mechanisms in the manipulator; and $m$ is the number of these mechanisms.

From previous considerations, the optimal problem to be solved takes the form of a constrained nonlinear programming problem, as follows:
\begin{equation}
	\hspace*{-0.1cm}\underset{\bld{\Xi}, \ \bld{c}}{\operatorname{minimize}}  \ \  f \ = \ \tfrac{1}{2} \Delta_t \sum_{t=t_0}^{t_M} \left( \sum_{i=1}^{n_{a}} \frac{f_{x_i} \cdot v_{x_i}} {\eta_{\text{EMLA}_i} (f_{x_i} , v_{x_i})} \right)^{2}
	\label{eq:costf2}
\end{equation}

subject to
\begin{subequations}
\begin{eqnarray}
        \left[ \bld{f}_{x}, \ \bld{v}_{x} \right] & = &  \bld{ID}\left( \bld{\theta}, \dot{\bld{\theta}}, \ddot{\bld{\theta}}, robot(\bld{\Xi}) \right) \label{const_a} \\[-2pt]
		\bld{x}_{r}(t) & = & \bld{G}_{tcp}(\bld{\theta}) \label{const_b} \\[-2pt]
		\dot{\bld{x}}_{r}(t) & = & \bld{J}(\bld{\theta})\dot{\bld{\theta}} \label{const_c} \\[-2pt]
		\ddot{\bld{x}}_{r}(t) & = & \dot{\bld{J}\,}(\bld{\theta},\dot{\bld{\theta}})\dot{\bld{\theta}} + \bld{J}(\bld{\theta})\ddot{\bld{\theta}} \label{const_d} \\[-2pt]
		\bld{v}_{x_{low}} &  \leq & \bld{v}_{x} \ \ \leq \ \ \bld{v}_{x_{up}} \label{const_e} \\[-2pt]
		\bld{f}_{x_{low}} &  \leq & \bld{f}_{x} \ \ \leq \ \ \bld{f}_{x_{up}} \label{const_f} \\[-2pt]
		0 &  \leq & \bld{\xi} \ \ \leq \ \ \bld{\xi}_{up} \label{const_g} \\[-2pt]
		\bld{\theta}_{low} &  \leq & \bld{\theta} \ \ \leq \ \ \bld{\theta}_{up} \label{const_h} \\[-2pt]
		\dot{\bld{\theta}}_{low} &  \leq & \dot{\bld{\theta}} \ \ \leq \ \ \dot{\bld{\theta}}_{up} \label{const_i} \\[-2pt]
		0 &  \leq & x(t) \ \ \leq \ \ L_c \label{const_j} \\[-2pt]
		L_{c0} + x(t) + L_c &  \leq & L + L_1 \label{const_k} \\[-2pt]
		q + q_1 + q_2 & = & -\pi \label{const_l} \\[-2pt]
		\bld{G}_{T1}\left( \bld{\theta} \right) & = & \bld{G}_{T2}\left( \bld{\theta} \right) \label{const_m}
	\end{eqnarray}
 \label{eq:costf2_}
 \end{subequations}
where $\Delta_t\defeq t_k-t_{k-1}$; the cost function $f$ represents the discrete time integral of the sum of the output power of EMLAS with $n_a$ as the number of linear actuators; $\eta_{\text{EMLA}}$ is the function defined in (\ref{system_efficiency}) which stands for EMLA efficiency; and $f_{x_i}$ and $v_{x_i}$ denote the scalar force and velocity of the $i$-th actuator, respectively. It should be noted that the cost function $f$ stands for the energy expenditure because it is the time integral of power. The variables $\bld{x}_{r}(t)$, $\dot{\bld{x}}_{r}(t)$, and $\ddot{\bld{x}}_{r}(t)$ represent the position, velocity, and acceleration of the TCP trajectory reference in the operational space, and $\bld{J}(\bld{\theta})$ and $\dot{\bld{J}\,}(\bld{\theta},\dot{\bld{\theta}})$ are the manipulator Jacobian and its time derivative, respectively. The actuator's velocity and force are bounded through (\ref{const_e}-\ref{const_f}). The upper and lower values are denoted by the subscripts $up$ and $low$, respectively. Positive lengths are forced with (\ref{const_g}) and joint limits with (\ref{const_h}-\ref{const_i}).

The trigonometric consistency for each closed kinematic chain in the robot is achieved via the constraints (\ref{const_j}-\ref{const_m}). For instance, (\ref{const_k}) forces the upper chain to be longer than the lower one to avoid singularities in (\ref{eq:qus}), while (\ref{const_l}) ensures the triangle property (\ref{eq:triangle}). The last constraint (\ref{const_m}) imposes the kinematic loop closure by forcing the upper and lower serial chains to end in the same $SE(3)$ element; see (\ref{eq:frames}).
\section{Numerical Simulations and Results}
\begin{figure}[h!] 
	\centering
	\includegraphics[trim={0.0cm 0.0cm 0.0cm 0.0cm},clip,width=7cm]{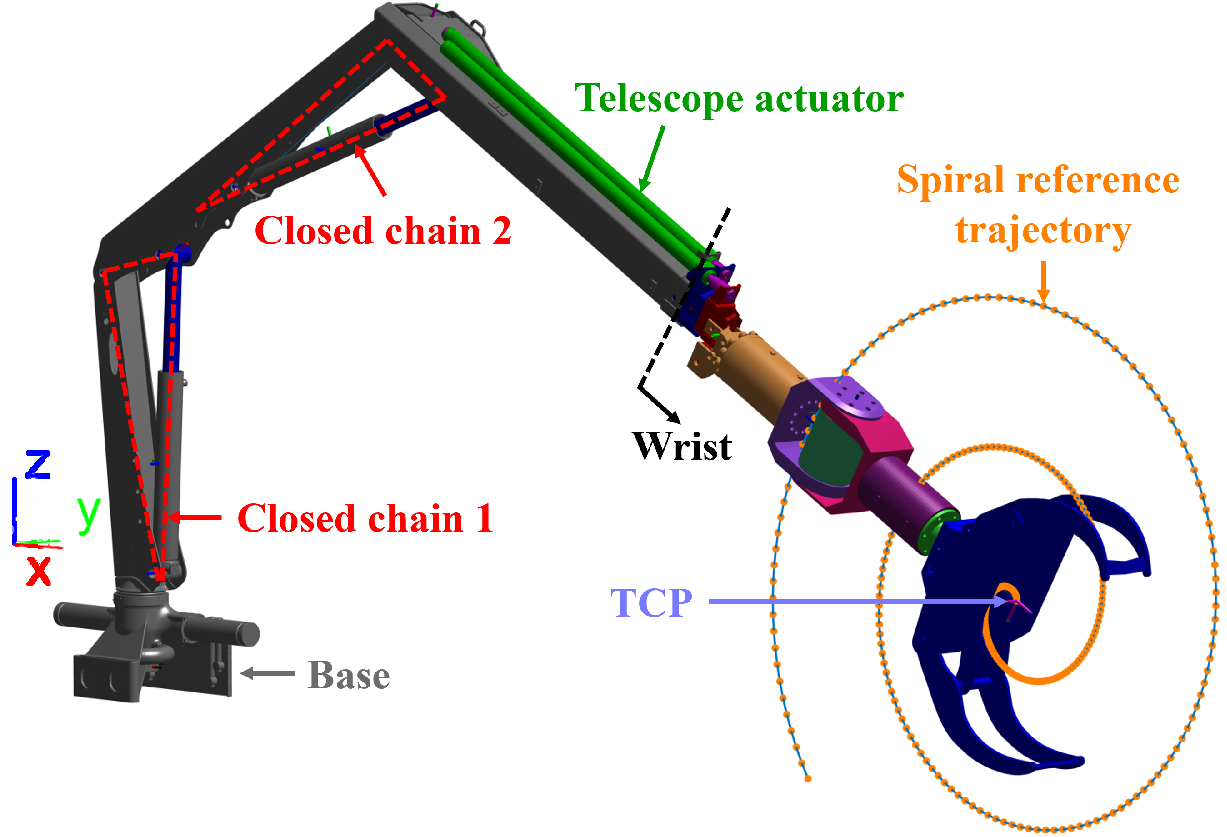}
	\vspace*{-0.1cm}
	\caption{Heavy-duty parallel-serial manipulator HIAB. It is a 7-DoF robot endowed with two parallel mechanisms. Its TCP is following a spiral reference trajectory that can cover a considerable area of the operational space, as its rotational base and spheric wrist are considered fixed.}
	\label{fig:hiab}
\end{figure}

\begin{figure}[h!] 
	\centering
	\includegraphics[trim={0cm 0.0cm 0.0cm 0cm},clip,width=7.5cm]{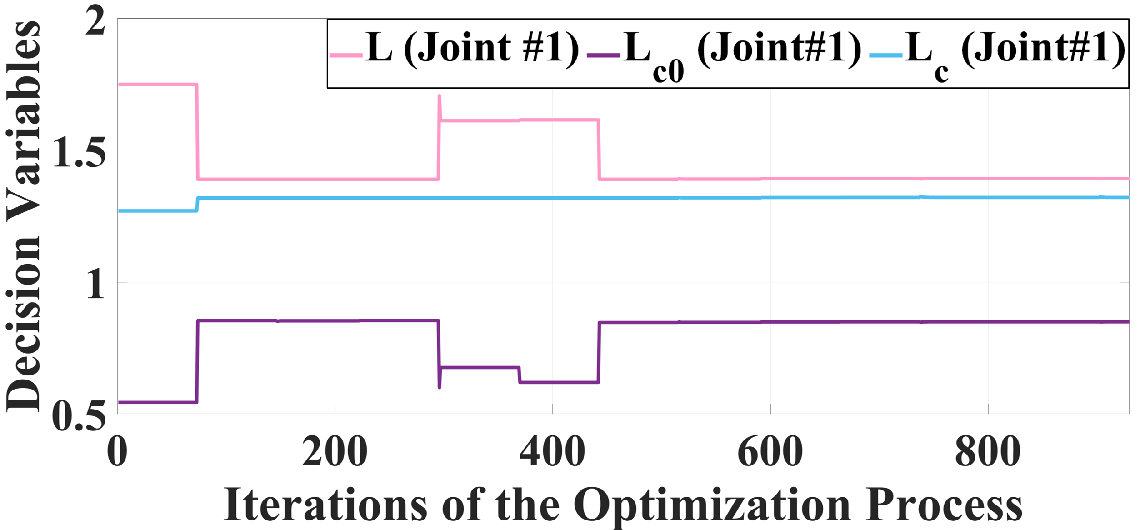}
	\vspace*{-0.1cm}
	\caption{ Optimization results: Variation in decision variables of first closed chain in the optimization process.}
	\label{fig:Decision_variables_1}
\end{figure}

\begin{figure}[h!] 
	\centering
	\includegraphics[trim={0cm 0.0cm 0.0cm 0cm},clip,width=7.5cm]{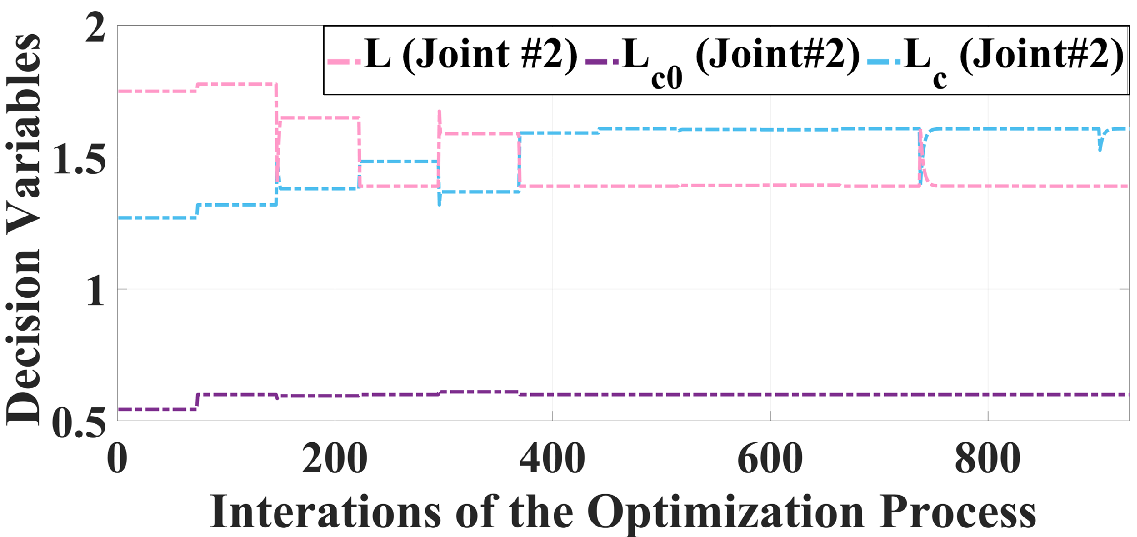}
	\vspace*{-0.1cm}
	\caption{ Optimization results: Variation in decision variables of second closed chain in the optimization process.}
	\label{fig:Decision_variables_2}
\end{figure}

\begin{figure}[h!] 
	\centering
	\includegraphics[trim={0.0cm 0.0cm 0.0cm 0.0cm},clip,width=9.2cm]{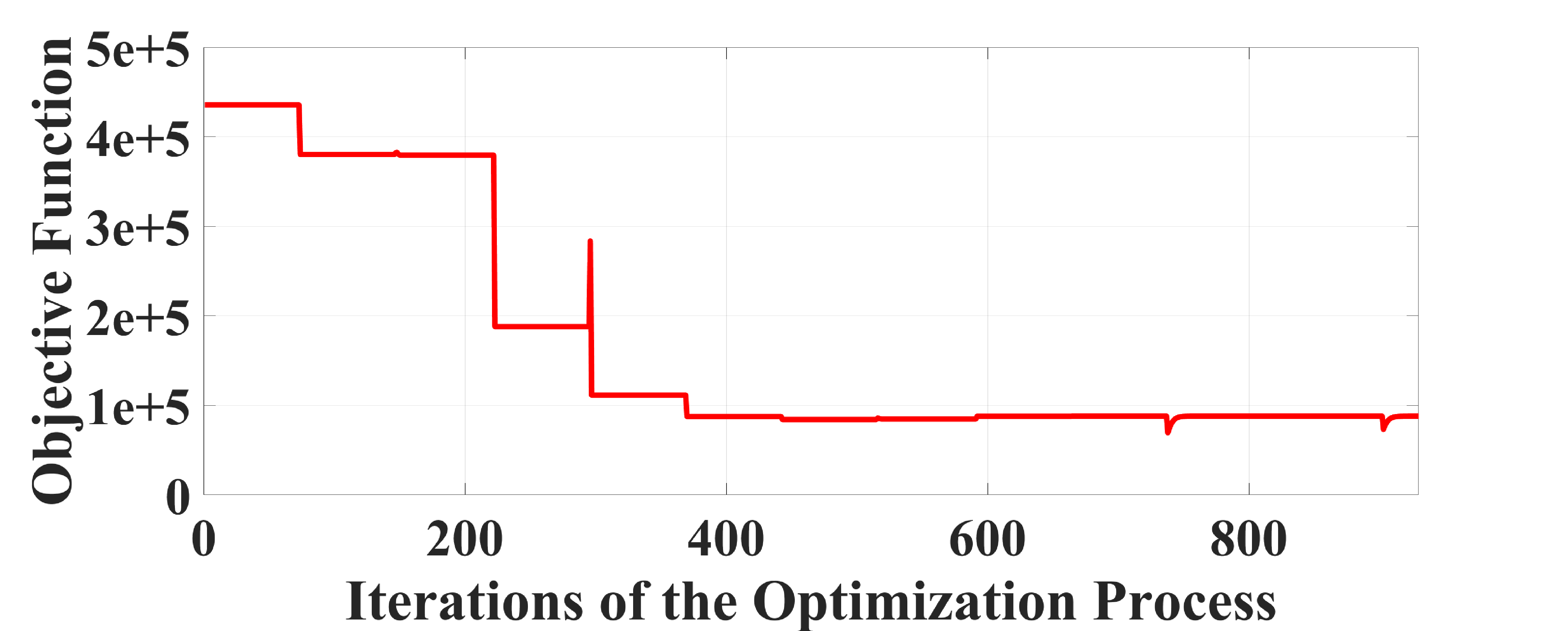}
	\vspace*{-0.1cm}
	\caption{Variation in the objective function during the optimization process. The amount of energy expenditure is minimized according to the expression (\ref{eq:costf2}) in around 900 iterations of SQP.}
	\label{fig:Objective_function}
\end{figure}

\begin{figure}[h!] 
	\centering
	\includegraphics[trim={0.0cm 0.0cm 0.0cm 0.0cm},clip,width=6.25cm]{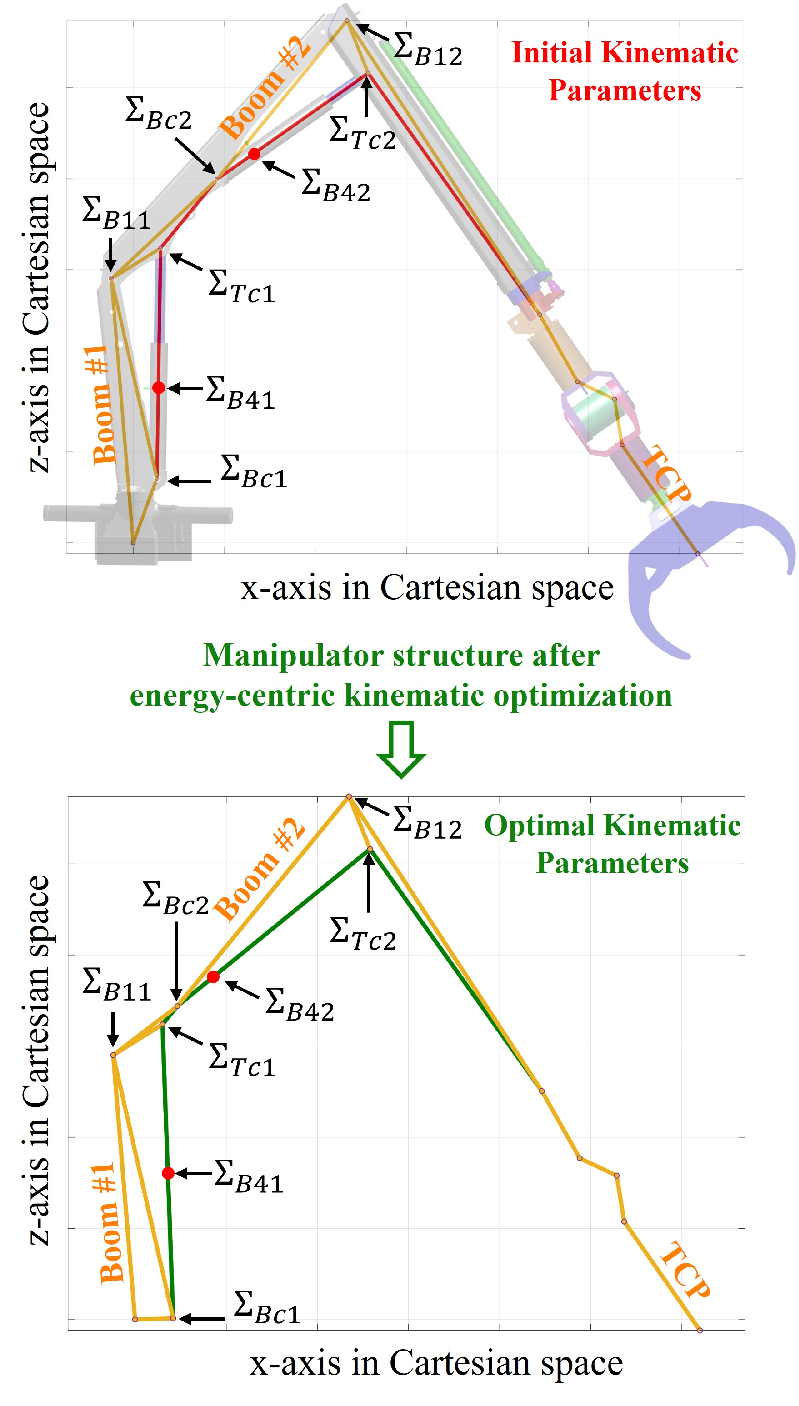}
	\vspace*{-0.1cm}
	\caption{Comparison among manipulator structures, which illustrates the primal solution (upper) and optimal solution (lower) when closed-loop kinematic parameters are optimized according to the optimal problem (\ref{eq:costf2}-\ref{eq:costf2_}).}
	\label{fig:Initial_Structure_Modified}
\end{figure}

Let us illustrate the kinematic parameter optimization problem described in the last section with a heavy-duty 7-DoF parallel-serial manipulator HIAB robot, as depicted in Fig. \ref{fig:hiab}. For this study, we analyze only the motion provided by the three linear actuators, which are both closed chains and telescope joint. This simulation was implemented in Matlab with \textit{fmincon} as the solver in a laptop endowed with standard computational capabilities, 1.80GHz, and 16Gb RAM. We set the BSpline parameters as $N = 22, M = 25, \ \mbox{and} \ \mathcal{T} = \{0 \ 0.25 \ \cdots \ 6.25\}$. Conversely, link lengths were initially set as $\bld{\xi}_1 = [1.75, 0.544, 1.27]$ for the first closed chain and $\bld{\xi}_2 = [1.75, 0.55, 1.2]$ for the second. A spiral reference trajectory in the X-Z plane was forced to track (see Fig. \ref{fig:hiab}) because this shape covers most of the operational space when only the closed-loop motion is enabled. Fig. \ref{fig:Decision_variables_1} and Fig. \ref{fig:Decision_variables_2} illustrate the variation in decision variables during the optimization process in the first and second closed chain accordingly, while Fig. \ref{fig:Objective_function} depicts the objective function value during the optimization iterations. To visualize the changes in the structure of the heavy-duty manipulator, Fig. \ref{fig:Initial_Structure_Modified} shows the topology of the manipulator with initial and optimal kinematic parameters.
%


\section{Conclusions}
This paper presents a comprehensive approach to integrating EMLAs into heavy-duty manipulator structures for sustainable OHM design. Focusing on the structural optimization of a prevalent closed kinematic chain configuration used in heavy-duty manipulators, we have developed an energy-centric optimization framework to minimize energy consumption. This study surpasses the limitations of previous studies that focused solely on structural or dynamic analysis by considering the efficiency of actuation into account. By introducing a high-fidelity analytical model of the full parallel-serial manipulator dynamics, we determined optimal kinematic parameters for harmonious EMLA integration. The proposed methodology holds significant potential to reduce the environmental impact of the transportation and logistics industry by facilitating the transition from ICEs to clean-tech electric actuators in heavy-duty BEVs. Future research can explore advanced control strategies, more sophisticated EMLA models, the harmonic response and natural frequency and an analytical representation of the whole working space.


	\bibliography{biblio}
	\bibliographystyle{unsrt}

\end{document}